\documentclass[letterpaper, 10 pt, conference]{ieeeconf}  % Comment this line out if you need a4paper

\IEEEoverridecommandlockouts                              % This command is only needed if 
\usepackage{cite}
\usepackage{csquotes}
\usepackage{amsmath,amssymb,amsfonts}
\usepackage{algorithmic}
\usepackage{graphicx}
\usepackage{textcomp}
\usepackage[dvipsnames]{xcolor}
\usepackage{balance}
\usepackage[hidelinks]{hyperref}
\usepackage{lipsum}
\usepackage{breqn}
\usepackage{booktabs}
\usepackage{siunitx}
\usepackage{adjustbox}
\usepackage{tikz}
\usetikzlibrary{calc,patterns,decorations.pathmorphing,decorations.markings,positioning,backgrounds,arrows.meta,shapes,fit,matrix,spy,shapes.geometric}
\usepackage{multirow}
\usepackage{booktabs}
\usepackage{colortbl}
\usepackage[caption=false, font=footnotesize]{subfig}

\def\E#1{\ensuremath{\text{E}\left \{#1 \right \}}}

\title{\LARGE \bf Towards Robotised Palpation for %Thyroid
Cancer Detection through Online Tissue Viscoelastic Characterisation with a Collaborative Robotic Arm
\thanks{We acknowledge the support of the MUR PNRR project FAIR - Future AI Research (PE00000013).}}

\author{Luca Beber$^{1,2}$, Edoardo Lamon$^{1,3}$, Giacomo Moretti$^{4}$, Daniele Fontanelli$^{4}$, Matteo Saveriano$^{4}$, Luigi Palopoli$^{1}$
\thanks{$^{1}$Department of Information Engineering and Computer Science, Universit\`a di Trento, Trento, Italy. \tt\small luca.beber@unitn.it}
\thanks{$^{2}$DRIM, Ph.D. of national interest in Robotics and Intelligent Machines.}
\thanks{$^{3}$Human-Robot Interfaces and Interaction, Istituto Italiano di Tecnologia, Genoa, Italy.}
\thanks{$^{4}$Department of Industrial Engineering, Universit\`a di Trento, Trento, Italy.}
}

\begin{document}

\maketitle
\thispagestyle{empty}
\pagestyle{empty}

%%%%%%%%%%%%%%%%%%%%%%%%%%%%%%%%%%%%%%%%%%%%%%%%%%%%%%%%%%%%%%%%%%%%%%%%%%%%%%%%
\begin{abstract}
%In the past two decades, RMSI exams have become increasingly popular. Of these, the palpation examination has received the most attention in robotics. 
%Performing a palpation examination at the appropriate time can be critical, but it is often delayed due to a shortage of time and medical personnel.

This paper introduces a new method for estimating the penetration of the end effector and the parameters of a soft body using a collaborative robotic arm. This is possible using the dimensionality reduction method that simplifies the Hunt-Crossley model. The parameters can be found without a force sensor thanks to the information of the robotic arm controller. To achieve an online estimation, an extended Kalman filter is employed, which embeds the contact dynamic model. The algorithm is tested with various types of silicone, including samples with hard intrusions to simulate cancerous cells within a soft tissue. The results indicate that this technique can accurately determine the parameters and estimate the penetration of the end effector into the soft body. These promising preliminary results demonstrate the potential for robots to serve as an effective tool for early-stage cancer diagnostics.
\end{abstract}

%%%%%%%%%%%%%%%%%%%%%%%%%%%%%%%%%%%%%%%%%%%%%%%%%%%%%%%%%%%%%%%%%%%%%%%%%%%%%%%%
\section{INTRODUCTION}
Palpation screening exams are noninvasive and inexpensive procedures that can help to detect cancer or other abnormalities at early stages when they are most treatable. During the exam, healthcare professionals use their hands to feel for lumps, masses, or enlarged lymph nodes that could be signs of cancer. However, the efficacy of the visit highly depends on the healthcare provider's experience and skill in feeling for abnormalities, especially when they are small, deep within the body, or located in soft tissues. Robotised palpation exams have the potential to reduce the subjectivity inherent in human performance by standardising the procedure. Therefore, this technology could improve the reliability of the results, and reduce the influence of the examiner's experience. 
% In addition, they might reduce the patient discomfort and anxiety caused in some cases by the human presence.
Another important benefit of robotised palpations could be the execution of the exams in geographical areas that are not covered by adequate health-care services.
%In addition, they could allow for examinations in remote areas or situations where a healthcare provider might not be readily available, holding promise for telemedicine applications, and enabling specialists to perform examinations remotely.

In the past decades, several robotic palpation systems have been introduced for the detection of hard bodies within soft tissues, leveraging artificial tactile sensors and sensorised probes~\cite{dario1988advanced,herzig2018variable,scimeca2020structuring}, and mechanical property characterisation with inverse finite element estimation methods~\cite{samur2007robotic,ahn20212robotic}.
A key element of any robotised palpation system is a solution for online estimation of the biological tissue parameters. Indeed, palpation exams require to locate stiffer points on the body~\cite{Zheng1996AnIn-vivo,Barbe2007InInsertions}. 
The same technology is also important for other medical applications. For example, in robot-assisted minimally invasive surgery (RMIS), haptic feedback can be implemented by accurately reconstructing the force on the end effector~\cite{Hashtrudi-Zaad1996AdaptiveTeleoperation}. Body-related information can also be applied to enable force-controlled navigation of a probe along the surface of the human body. In this case, it is important to ensure adequate levels of safety by regulating the force in contact with rigid body parts and accidental motions of the patient. ~\cite{love2004force,cortesao2006real,Beber2024AUltrasonography}.
% Today's Robot-assisted minimally invasive surgery (RMSI) currently lacks tactile feedback. This limits the surgeon’s ability to efficiently locate cancerous tissue, which can be up to 100 times stiffer than normal tissue~\cite{Weienborn2016DeformationMethods}. To overcome this limitation, haptic feedback can be implemented by accurately reconstructing the force on the end-effector. For this reason, in recent years many studies have focused on estimating the soft tissue parameters using collaborative robotic arms.

Different models suitable for estimation can be adopted to
characterise soft tissues. Pappalardo et
al.~\cite{Pappalardo2016Hunt-CrossleySurgery} showed that linear
models, such as the Kelvin-Voigt or Maxell models, have limitations as
they do not account for the geometry of the contact surface between
the end effector and the soft body. A better-suited choice is the
nonlinear Hunt-Crossley (HC) model, which
describes the tridimensionality of the contact and reconstructs the
force on the end effector with higher precision~\cite{hunt1975coefficient}. Such models are used
in combination with different estimation methods, such as non-linear
square
regression~\cite{diolaiti2005contact,Schindeler2018OnlineLinearization}
and Kalman
filters~\cite{Pappalardo2016Hunt-CrossleySurgery,Zhu2021ExtendedModel,Roveda2022SensorlessEstimation,Zhu2023IterativeIdentification}.

Despite their promising results, the aforementioned approaches share
two common limitations. The first is that the penetration depth is
assumed known~\cite{Zhu2021ExtendedModel}, which is unrealistic in
real-life applications. During a palpation test, there is hardly any
way to measure penetration. To overcome this problem, Roveda et
al.~\cite{Roveda2022SensorlessEstimation} propose an approach phase in
which the end effector is moved slowly until contact with the body
surface occurs. This solution is applicable within experimental
settings, where the surface of the palpated object remains constant
over time. However, this situation is very unlikely for medical
examinations. For example, in the palpation of a person's abdomen, its
surface is regularly inflated and deflated while breathing. The
second limitation arises from the reliance on highly precise force
sensors for the estimation. Such devices are expensive and prone to damage. Besides, they require a regular calibration,
which reduces the usability of the device.

%Another limitation is associated with the adoption of Hunt-Crossley model. The unknown exponential term causes strong nonlinearity in the model, and the estimation filter cover es slowly to the correct parameters and often highly complex filters are implemented to overcome this limitation.
   
To overcome the limitations above, we propose in this paper a method
capable of estimating at the same time the penetration inside the body
and the parameters of the model. This is possible thanks to the use of
the dimensionality reduction method that estimates the exponential term of the Hunt-Crossley method
knowing the shape of the end effector~\cite{Popov2015MethodFriction}. In addition, we introduce a
dynamical model that links the dynamics of the robotic arm, the end
effector position, and the soft surface mechanical behaviour. The
combined use of dimensionality reduction and dynamical model allows us
to: 1. find the amount of penetration ensuring a fast convergence of
the model parameters, 2. avoid direct force measurements.

The method is evaluated with a robotic arm equipped with a spherical end effector and four different silicone samples, two of which contained stiffer material to simulate the presence of a foreign mass. The experiments objective is to estimate in real time the viscoelastic model parameters of the specimen and the correct amount of penetration. Moreover, the force reconstructed with the estimated data has been compared with the one registered by the force torque sensor.
Our results show that the package of solutions proposed in the paper reduces the complexity of the system and enhances its physical robustness without any negative impact on the accuracy of the measurements.

% The paper is organised as follows: the second section describes the methods used in our work; in section 3, we describe the experimental setup and the materials used; section 3, is devoted to the analysis of the experimental results, and finally, in section 4, we discuss the results and possible future works.

\section{METHODS}

% \subsection{Background on Contact Models}
In this section, first the contact models used to characterise the soft body will be presented; then two different estimation strategies (with and without force sensing) will be illustrated. Finally we will briefly summarise the use of the extended Kalman filter. 
One of the simplest way to describe the interaction between a hard indenter and a soft deformable body is the Kelvin-Voigt (KV) model. This model is popular because, despite its simplicity, it describes with good approximation the dynamics of the contact. The soft body is treated as an ideal viscoelastic material and contact is considered punctual. The force generated by the soft body is modelled by the combination of a linear spring and a damper acting in parallel, and it can be written as
\begin{equation}
        F_{M}(d) = 
        \begin{cases}
                k_M d + c_M \dot{d}, \;\; & d \geq 0, \\
                0 \;\; & d < 0,
        \end{cases}
        \label{eq:kv_force}
\end{equation}
where $F_{M}(d)$ is the force generated by the material, $d$ is the penetration inside the body, $\dot{d}$ is the velocity of penetration, $k_M$ is the stiffness of the body and $c_M$ is the damping. Several studies have shown that
%, %for medical robotics applications, the limits of this model can not be accepted and
%in case of small penetration, 
the model is inaccurate since it does not capture 
important nonlinear effects when the contact surface
cannot be reduced to a point. 
For this reason, nonlinear models such as the Hunt-Crossley (HC) model are preferable~\cite{diolaiti2005contact,Pappalardo2016Hunt-CrossleySurgery}. The HC model is energetically correct %with what happens in reality 
and exposes an explicit dependence between the damping term and the penetration depth. 
However, the presence of an unknown exponential term makes the estimation very difficult.
%The reason that led us to consider another model is that the exponential term is unknown and has to be estimated, which is not an easy task and increases the nonlinearity of the system. 

\begin{figure}[t]
        \centering
        \includegraphics{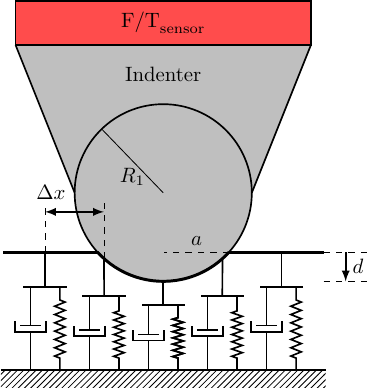}
        \caption{Schematic of a spherical indenter in constant with a viscoelastic half-plane. $\Delta x$ is the distance between the viscous elements, $a$ represents half of the projection of the circle's portion in contact, $d$ is the maximum penetration and $R_1$ is the equivalent radius of the sphere in the viscoelastic halfplane.}
        \label{fig:schematic}
\end{figure}

If the shape of the end effector is known, the contact can be
described using the dimensionality reduction method
(DRM)~\cite{Popov2015MethodFriction}. In our previous work, we showed
that using a robotic arm and an indenter of known shape, the static
force can be reconstructed
precisely~\cite{Beber2024ElasticityArm}. In this work, the model was modified to include the dynamic response of the material. The idea behind the DRM
is that the forces generated by a three-dimensional contact between an
axial symmetric indenter and a surface can be computed using the 2D
projection onto a plane.  For instance, the contact between a sphere,
with radius $R$, and a surface can be studied by limiting the analysis
to the contact between a circular arc, with radius $R_1 = R/2$, and a
set of parallel mass-damper elements spaced out of a tiny quantity
$\Delta x$. The deformation of each element is dependent on the
contact point between the indenter and the material. In the case of a
spherical indenter, the central elements are deformed more than the
elements on the borders, as shown in~\autoref{fig:schematic}.
In this study, we will focus on incompressible materials (Poisson ratio $\nu = 0.5$), i.e. materials for which the volume does not change under an external pressure's action. Such materials include biological tissues but also silicons~\cite{Sacks2003}. 
The viscoelastic formulation with DRM can be divided into two subproblems: the purely elastic case and the purely viscous case. In the purely elastic case, analysed in~\cite{Beber2024ElasticityArm}, the contribution of a single spring is 
\begin{equation}
        f_{M,i} = 4 G d_{i}\Delta x, 
        \label{eq:elastic_spring}
\end{equation}
where $f_{M,i}$ is the contribution of the $i-$th spring, $d_i$ is the
local penetration in the body, and $G$ is the shear elastic
modulus. Similarly, the contribution of a single damper can be found
by substituting position with velocity and elasticity with
 viscosity, i.e., in~\eqref{eq:elastic_spring}
$d\rightarrow \dot d$ and $G\rightarrow \eta$, thus obtaining
\begin{equation}
    f_{M,i} = 4 \eta \Dot d_{i} \Delta x,
    \label{eq:elastic_visc}
\end{equation}
where $\Dot d_{i}$ is the velocity of penetration and $\eta$ is the viscosity of the material. Integrating the sum of~\eqref{eq:elastic_spring} and~\eqref{eq:elastic_visc} with $\Delta x \rightarrow 0$
and approximating the circular arc by a parabola (as per the Hertzian theory), we obtain that the force action generated by the material on the end effector is given by
\begin{equation}
        F_M(d) =
        \begin{cases}
                \kappa d^{\frac{3}{2}} + \lambda   d^{\frac{1}{2}}
                \dot{d}, \;\; & d\geq0 ,\\
                0 \;\; & d<0 ,
        \end{cases}
        \label{eq:final_ext_contact}
\end{equation}
where the new model parameters $\kappa $ and $\lambda $ for an incompressible viscoelastic material are
\begin{equation}
  \kappa  = \frac{16 G}{3} \sqrt{R} \mbox{ and } \lambda  = 8\eta \sqrt{R}.
\end{equation}
Using~\eqref{eq:final_ext_contact}, it is therefore possible to model
nonlinear effects in the contact without having to estimate the
material-specific value of the exponent of the HC model. The complete
derivation can be found in~\cite{Popov2015MethodFriction,Beber2024ElasticityArm}.

% \begin{figure}[t]
% \centering
%         \input{figure/indentation.pdf}
%         \caption{Model of the system.}
%         \label{fig:indenter_DR}
% \end{figure}

\subsection{Online Estimation with Measured Force Inputs}
\label{sec:forceSensor}
The force measured by a force sensor positioned on the robot's end
effector can be written as the sum of the force generated by the soft
tissue and the inertia of the indenter. For the sake of simplicity, we will explain the formulation first with the Kelvin-Voigt model, and then adapt it with the DRM method.
Using the simpler KV model~\eqref{eq:kv_force}, the force on the end effector in contact with a soft surface can be written as
\begin{equation}
        F_{FT} = k_M d + c_M \dot{d}+ m_I \ddot{d},
        \label{eq:force_lin}
\end{equation}
where $m_I$ is the mass of the indenter attached to the force sensor. Solving~\eqref{eq:force_lin} for $\Ddot{d}$, we obtain the differential equation
\begin{equation*}
    \Ddot{d} = \frac{1}{m_I} \left( F_{FT} - k_M d - c_M
      \dot{d}\right) ,
\end{equation*}
from which it is easy to derive a state space representation of the dynamical system. The state vector $\boldsymbol{x}\in\mathbb{R}^4$ is defined,
which contains the penetration inside the body, the velocity of the
indenter, the stiffness, and the damping of the soft tissue,
$\boldsymbol{x} = \left[d, \dot d, k_M, c_M\right]^{\top}$. The corresponding discrete-time model,
$\boldsymbol{x}_{t+1}=f_t(\boldsymbol{x}_t,u_t,\boldsymbol{w}_t)$, assuming $\Delta T$ as sampling time, is
\begin{equation}
  \begin{aligned}
    x_{1,t+1}&= x_{1,t} + \Delta T x_{2,t} + w_1,\\
    x_{2,t+1}&= x_{2,t} + \dfrac{\Delta T}{m_I} \left(u_t -x_{1,t} x_{3,t} - x_{2,t}x_{4,t}\right) + w_{2},\\
    x_{3,t+1}&= x_{3,t},\\
    x_{4,t+1}&= x_{4,t}.
  \end{aligned}
  \label{eq:state_space_lin_FT}
\end{equation}
The input of the system $u_t$ is the force $F_{FT}$ measured by the
F/T sensor, $\boldsymbol{x}_t$ is the state at the current time step,
$\boldsymbol{x}_{t+1}$ is the state at the following time step and
$\boldsymbol{w}_t$ is the model noise.  It can be observed
from~\eqref{eq:state_space_lin_FT} that the stiffness and damping
coefficients are time-invariant, so no model uncertainty is considered
in their dynamics. Moreover, since the integration time interval
$\Delta T$ is very small, we assumed the velocity to be constant in
$\Delta T$.  Furthermore, since the robot end effector is always in
contact during the estimation phase, the velocity of the indenter
inside the soft body is the same as the velocity of the end
effector. The quantity can be calculated by utilising the velocity of
the robot's joint and the Jacobian matrix through direct differential
kinematics. The measurement function
$z_t = h_t(\boldsymbol{x}_t,v_t)$, instead, can be defined
\begin{equation}
        z_t = x_{2,t} + v_t,
        \label{eq:measurament}
\end{equation}
where $v_t$ is the measurement uncertainty.

The same procedure can be followed substituting the nonlinear model
in~\eqref{eq:final_ext_contact} in~\eqref{eq:force_lin}, obtaining a
new update rule for the velocity of the end effector. The state vector is in this case  $\boldsymbol{x} = \left[d, \dot d, \kappa , \lambda
\right]$, and the velocity update rule is 
\begin{equation}
  x_{2,t+1} = x_{2,t} + \frac{\Delta T}{m_I} \left(u_t -x_{1,t}^{\frac{3}{2}} x_{3,t} - x_{1,t}^{\frac{1}{2}}x_{2,t}x_{4,t}\right) + w_{2}.
        \label{eq:state_space_NL_FT}
\end{equation}

\subsection{Online Sensorless Estimation with Force Approximation}
% In the previous section, we discussed how the force acting on the end effector is related to the control of the robot.
The following section presents a method for estimating the soft tissue
parameters without a 6-axis force sensor.  The method is underlined by
the observation that the interaction force of the robot with the
environment, generated by a Cartesian impedance controller, is in
general equal to the external forces measured by the F/T
sensor. Hence, in the above-mentioned conditions, it is possible to
exploit a Cartesian impedance controller to estimate the interaction
forces without having a force torque sensor mounted on the end
effector. A similar approach has been proposed by Roveda et
al. in static conditions~\cite{Roveda2022SensorlessEstimation}. However, in this approach, the impedance control
law is equated to the force generated by a spring neglecting the damping term and the equilibrium position is evaluated prior to the elastic estimation with a strict procedure which requires the robot to be positioned in close proximity to the contact surface.

In a Cartesian impedance controller, the physical interaction of the end-effector with the environment $\boldsymbol{F}^{ext}_{ee}$ approximates a mass-spring-damper system with desired parameters
\begin{equation}
\boldsymbol{\Lambda}_d\ddot{\Tilde{\boldsymbol{x}}}_{ee} +
\boldsymbol{D}_d\dot{\Tilde{\boldsymbol{x}}}_{ee} +
\boldsymbol{K}_d\Tilde{\boldsymbol{x}}_{ee} =
\boldsymbol{F}^{ext}_{ee} ,
    \label{eq:cartesian_impedance}
\end{equation}
where
$\Tilde{\boldsymbol{x}}_{ee} = \boldsymbol{x}_d - \boldsymbol{x}_{ee}
\in\mathbb{R}^{m}$ is the Cartesian position error $\boldsymbol{x}_d$,
$\boldsymbol{\Lambda}_d \in\mathbb{R}^{m \times m}$,
$\boldsymbol{D}_d \in\mathbb{R}^{m \times m}$ and
$\boldsymbol{K}_d \in\mathbb{R}^{m \times m}$ are the desired
Cartesian inertia, damping, and stiffness, respectively. In general,
the implementation of a closed-loop scheme to render the behaviour in
\eqref{eq:cartesian_impedance} requires a force/torque sensor
capable of measuring $\boldsymbol{F}^{ext}_{ee}$. However, under the
assumption of natural inertia
$\boldsymbol{\Lambda}_d = \boldsymbol{\Lambda}(\boldsymbol{x})$, the
feedback of external forces can be
avoided~\cite{dietrich2021practical}. The control law
in~\eqref{eq:cartesian_impedance}
% further simplifies in the case of equilibrium points, where $\ddot{\boldsymbol{x}}_d=\boldsymbol{0}$ and $\dot{\boldsymbol{x}}_d=\boldsymbol{0}$, as follows:
can be reformulated as
\begin{equation}
     \boldsymbol{\Lambda}(\boldsymbol{x}) \Ddot{\Tilde{{\boldsymbol{x}}}}_{ee} + \boldsymbol{D}_d \Dot{\Tilde{{\boldsymbol{x}}}}_{ee} + \boldsymbol{K}_d\Tilde{\boldsymbol{x}} = \boldsymbol{F}^{ext}_{ee}.
    \label{eq:cartesian_impedance_simplified}
\end{equation}

As discussed in Section~\ref{sec:forceSensor}, also in this case we
will start by formulating the estimation model using the linear KV
model.
%and then we will show the adaptation required when using the  
Since the force measured by the F/T sensor is equivalent to the force
generated by the impedance controller, to estimate the force acting on
the material, we can equate~\eqref{eq:force_lin}
with~\eqref{eq:cartesian_impedance_simplified} in the contact
direction.  Assuming that the contact occurs in the $z$-direction, we
can extract an equation in such a direction from the general
equation~\eqref{eq:cartesian_impedance_simplified}. Given
$\boldsymbol{\Lambda}(x) = \left[ \Lambda_{ij}\right]$,
$\boldsymbol{D}_{d}= \left[ D_{ij} \right]$ and
$\boldsymbol{K}_{d}= \left[ K_{ij} \right]$, the resulting equation
is as follows
\begin{equation}
        \Lambda_{33} \ddot{\tilde{z}} +  D_{33} \dot{\tilde{z}} + K_{33} \tilde{z} = k_M d + c_M \dot{d}+ m_I \ddot{d}.
        \label{eq:force_lin1}
\end{equation}
As the robot end effector is in contact with the material, we have $\dot{z}_e = -\dot{d}$ and $\ddot{z}_e = -\ddot{d}$. Using this information we can rewrite~\eqref{eq:force_lin1} as
\begin{equation}
        \Lambda_{33} (\ddot{z}_d+\ddot{d}) +  D_{33}  (\dot{z}_d + \dot{d}) + K_{33} \Tilde{z} + k_M d + c_M \dot{d}+ m_I \ddot{d} = 0,
        \label{eq:force_lin2}
\end{equation}
and, finally, solving~\eqref{eq:force_lin2} for $\ddot{d}$ to obtain
\begin{dmath}
  \ddot{d} = -\frac{1}{m_I + \Lambda_{33}} \left(\Lambda_{33}
    \ddot{z}_d + D_{33} \dot{z}_d + K_{33} \tilde{z} + (D_{33} +
    c_M) \dot{d} + k_M d\right).\label{eq:force_lin3}
\end{dmath}
Using~\eqref{eq:force_lin3} we can write the new discrete-time model
of the
system. %The input becomes three dimensional, $\boldsymbol{u}\in\mathbb{R}^3$, and contains
The input now becomes
$ \boldsymbol{u} = \left[ \tilde{z}, \dot{z}_d, \ddot{z}_d,
  \Lambda_{33} \right]^T \in\mathbb{R}^4$.  The new discrete time
model
$\boldsymbol{x}_{t+1}=f(\boldsymbol{x}_t,\boldsymbol{u}_t,\boldsymbol{w}_t)$
is the same as in~\eqref{eq:state_space_lin_FT} except for the
definition of $x_{2,t+1}$, redefined as
\begin{dmath}
        x_{2,t+1} = x_{2,t} - \frac{\Delta T}{m_I+u_{4,t}} (u_{4,t}
        u_{3,t} + D_{33} u_{2,t} + K_{33} u_{1,t} + x_{3,t} x_{1,t}
        + (D_{33} + x_{4,t})x_{2,t} ) + w_{2,t}. \label{eq:im_sys}
\end{dmath}
Similarly, the measurement function~\eqref{eq:measurament} remains the same.

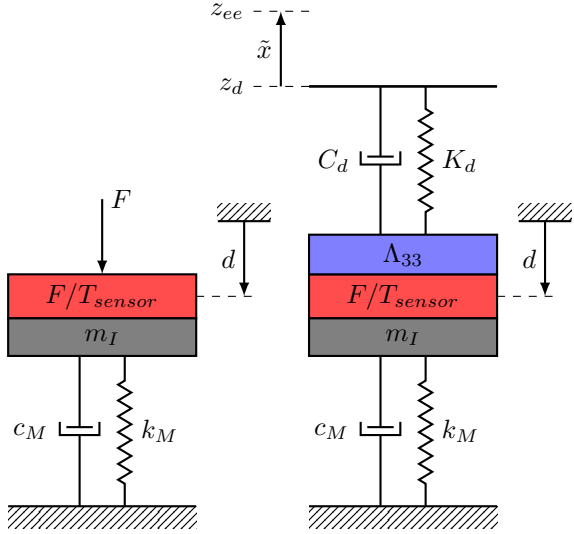
\begin{figure}[t!]
        \centering
        \begin{tikzpicture}[every node/.style={outer sep=0pt},thick,
    mass/.style={draw,thick},
    spring/.style={thick,decorate,decoration={zigzag,pre length=0.3cm,post
    length=0.3cm,segment length=6}},
    spring2/.style={thick,decorate,decoration={zigzag,pre length=0.7cm,post
    length=0.7cm,segment length=6}},
    ground/.style={fill,pattern=north east lines,draw=none,minimum
    width=0.75cm,minimum height=0.3cm},
    dampic/.pic={\fill[white] (-0.1,-0.3) rectangle (0.3,0.3);
    \draw (-0.3,0.3) -| (0.3,-0.3) -- (-0.3,-0.3);
    \draw[line width=1mm] (-0.1,-0.3) -- (-0.1,0.3);}]

    \tikzstyle{damper}=[thick,decoration={markings,  
    mark connection node=dmp,
    mark=at position 0.5 with 
    {
    \node (dmp) [thick,inner sep=0pt,transform shape,rotate=-90,minimum width=15pt,minimum height=3pt,draw=none] {};
    \draw [thick] ($(dmp.north east)+(2pt,0)$) -- (dmp.south east) -- (dmp.south west) -- ($(dmp.north west)+(2pt,0)$);
    \draw [thick] ($(dmp.north)+(0,-5pt)$) -- ($(dmp.north)+(0,5pt)$);
    }
    }, decorate]

    \node[mass,minimum width=2.5cm,minimum height=0.5cm,fill=red!70, align=center,anchor=center] (ft_sensor-) {$F/T_{sensor}$};
    \node[mass,minimum width=2.5cm,minimum height=0.5cm,fill=black!50,align=center,anchor=north east] (indenter-) at (ft_sensor-.south east)  {$m_I$};
    \node[below=2.5cm of ft_sensor-, ground, minimum width=2.5cm, minimum height=3mm, anchor=north] (g1-){};
    % \node[above=2.5cm of ft_sensor, draw, line width=1pt, minimum width=2.5cm, minimum height=0pt, inner sep=0pt, anchor=north] (line){};
    \draw (g1-.north west) -- (g1-.north east);

    \draw[spring] ([xshift=3mm]g1-.north) coordinate(aux) 
    -- (aux|-indenter-.south) node[midway,right=1mm]{$k_M$};

    \draw[damper] ([xshift=-3mm]g1-.north) coordinate(aux')
    -- (aux'|-indenter-.south)node[midway,left=3mm]{$c_M$};

    % \draw[thin] (ft_sensor.east) -- ++ (1,0) coordinate[midway](aux_ft);
    \draw[latex-] (ft_sensor-.north) -- ++ (0,1) node[right]{$F$};
    % \draw[thin,dashed] (line.west) -- ++ (-0.75,0) coordinate[midway](aux_line) node[left]{$x_d$};
    % \draw[-latex] (aux_line) -- ++(0,1)  node[midway,left]{$\Tilde{x}$} node[dashed](xt){}; 
    % \draw[thin,dashed, yshift=1cm] (line.west|-xt) -- ++ (-0.75,0) node[left]{$z_{ee}$};
    \draw[thin,dashed] (ft_sensor-.east) -- ++ (0.75,0) coordinate[pos=0.85](aux_ft1'-);
    \draw[latex-] (aux_ft1'-) -- ++ (0,1) node[midway,left]{$d$}
    node[above,ground,minimum height=1mm,minimum width=7mm] (g'-){};
    \draw[thick] (g'-.south west) -- (g'-.south east);

    \node[mass,minimum width=2.5cm,minimum height=0.5cm,fill=red!70, align=center,anchor=center,xshift=4cm] (ft_sensor) {$F/T_{sensor}$};
    \node[mass,minimum width=2.5cm,minimum height=0.5cm,fill=black!50,align=center,anchor=north east] (indenter) at (ft_sensor.south east)  {$m_I$};
    \node[mass,minimum width=2.5cm,minimum height=0.5cm,fill=blue!50,align=center,anchor=south east] (lambda) at (ft_sensor.north east)  {$\Lambda_{33}$};
    \node[below=2.5cm of ft_sensor, ground, minimum width=2.5cm, minimum height=3mm, anchor=north] (g1){};
    \node[above=2.5cm of ft_sensor, draw, line width=1pt, minimum width=2.5cm, minimum height=0pt, inner sep=0pt, anchor=north] (line){};
    \draw (g1.north west) -- (g1.north east);

    \draw[spring] ([xshift=3mm]g1.north) coordinate(aux) 
    -- (aux|-indenter.south) node[midway,right=1mm]{$k_M$};
    \draw[spring] ([xshift=3mm]line.north) coordinate(aux)
    -- (aux|-lambda.north) node[midway,right=1mm]{$K_d$};

    \draw[damper] ([xshift=-3mm]g1.north) coordinate(aux')
    -- (aux'|-indenter.south)node[midway,left=3mm]{$c_M$};

    \draw[damper] ([xshift=-3mm]lambda.north) coordinate(aux')
    -- (aux'|-line.north)node[midway,left=3mm]{$C_d$};

    % \draw[thin] (ft_sensor.east) -- ++ (1,0) coordinate[midway](aux_ft);
    % \draw[latex-] (aux_ft) -- ++ (0,-0.5) node[right]{$F_M$};
    \draw[thin,dashed] (line.west) -- ++ (-0.75,0) coordinate[midway](aux_line) node[left]{$z_d$};
    \draw[-latex] (aux_line) -- ++(0,1)  node[midway,left]{$\Tilde{x}$} node[dashed](xt){}; 
    \draw[thin,dashed, yshift=1cm] (line.west|-xt) -- ++ (-0.75,0) node[left]{$z_{ee}$};
    \draw[thin,dashed] (ft_sensor.east) -- ++ (0.75,0) coordinate[pos=0.85](aux_ft1');
    \draw[latex-] (aux_ft1') -- ++ (0,1) node[midway,left]{$d$}
    node[above,ground,minimum height=1mm,minimum width=7mm] (g'){};
    \draw[thick] (g'.south west) -- (g'.south east);

\end{tikzpicture} 
        \caption{The left model represents the system described in~\eqref{eq:state_space_lin_FT}, where it is assumed that a force is acting on the F/T sensor. The sensor is connected to a mass, which is in turn connected to a spring-damper element. The model described in~\eqref{eq:im_sys} is shown on the right, where the impedance control model is displayed instead of the external force.}
        \label{fig:models}
\end{figure}

The model can be interpreted as two masses, each connected to a
spring-damper set, and rigidly connected to one another, as shown
in~\autoref{fig:models}. The first spring-damper element represents
the force generated by the impedance controller, with the desired
position as input. The mass of the impedance controller and the
indenter are divided by the force sensor, which measures all the
forces acting on the end effector. Finally, the second spring-damper
element represents the force generated by the soft tissue.

Again, the update of the penetration velocity~\eqref{eq:im_sys} can be rewritten using the DRM in~\eqref{eq:final_ext_contact} as
\begin{dmath}
        x_{2,t+1} = x_{2,t} - \frac{\Delta T}{m_I+u_{4,t}} (u_{4,t} u_{3,t} + D_{33} u_{2,t} + K_{33} u_{1,t} + x_{1,t}^{\frac{3}{2}} x_{3,t} + x_{1,t}^{\frac{1}{2}} x_{2,t} x_{4,t} + D_{33} x_{2,t} ) + w_{2,t}.\label{eq:im_sys_nl}
\end{dmath}

\subsection{Extended Kalman Filter}
The system unknown variables (both static and dynamic) can be estimated using an extended Kalman filter (EKF). This filter corrects the update of a given model using the information provided by a measurement. 
%^In practice, it is the non-linear version of the Kalman filter, where the system is linearized at the first order at each step. 
Each step of the filter is composed of two phases: the update phase and the correction phase. In the update phase, the previous values are used to compute the new ones
\begin{align*}
  & \boldsymbol{\Hat{x}}_{t+1}^- =
    f(\Hat{\boldsymbol{x}}_t,\boldsymbol{u}_t,\boldsymbol{w}_t) , \\
  & \boldsymbol{P}_{t+1}^- =
    \boldsymbol{A}_t\boldsymbol{P}_t\boldsymbol{A}^T_t
    +\boldsymbol{G}_t\boldsymbol{Q}_t\boldsymbol{G}^T_t ,
\end{align*}
where $\boldsymbol{A_t}$ and $\boldsymbol{G_t}$ are respectively the
gradient of $f$ with respect to $\boldsymbol{x}$ and $\boldsymbol{w}$
computed in
$\left[\Hat{\boldsymbol{x}}_t,\boldsymbol{u}_t,\E{\boldsymbol{w}_t}\right]$,
where $\E{\boldsymbol{w}_t} = 0$ is the expected value (mean) of the
uncertainties. $\boldsymbol{Q}_t$ is the possibly time varying
covariance matrix of the uncertainties $\boldsymbol{w}$, while
$\boldsymbol{P}_{t}$ is the customary covariance matrix of the
estimation error. For the correction step, we have
\begin{align*}
    & \boldsymbol{S}_{t+1} = \boldsymbol{H}_{t+1}\boldsymbol{P}_{t+1}^-\boldsymbol{H}_{t+1}^T + \boldsymbol{R}_{t+1},\\
    & \boldsymbol{W}_{t+1} = \boldsymbol{P}_{t+1}^- \boldsymbol{H}_{t+1}^T \boldsymbol{S}_{t+1}^{-1},\\
    & \Hat{\boldsymbol{x}}_{t+1} = \Hat{\boldsymbol{x}}_{t+1}^-
      \boldsymbol{W}_{t+1} (\boldsymbol{x}_{t+1} -
      h(\Hat{\boldsymbol{x}}_{t+1}^-,v_{t+1})) ,\\
    & \boldsymbol{P}_{t+1} = (\boldsymbol{I} -
      \boldsymbol{W}_{t+1}\boldsymbol{H}_{t+1})\boldsymbol{P}_{t+1}^-, 
\end{align*}
where $\boldsymbol{H}_{t+1}$ is the gradient of $h$ in respect to $x$
computed in $\left[\Hat{\boldsymbol{x}}_{t+1}^-, \E{v_{t+1}}\right]$,
where, again, $\E{v_{t+1}} = 0$. Finally, $\boldsymbol{R}_{t+1}$ is
the possibly time varying covariance matrix of the additive
uncertainties $v_{t+1}$.

\section{EXPERIMENTS}
%\subsection{Experimental Setup}
The feasibility of the estimation algorithm thus described for the
penetration, stiffness, and damping of soft bodies made of different
silicone materials is here discussed. To this end, a sinusoidal
desired trajectory is imposed on the impedance controller to obtain
the maximum information from the palpation,
\begin{equation*}
    z_d = z_0 + z_1 \sin(4 \pi ) + z_2 \sin(8 \pi).
\end{equation*}

\begin{figure*}[t!]
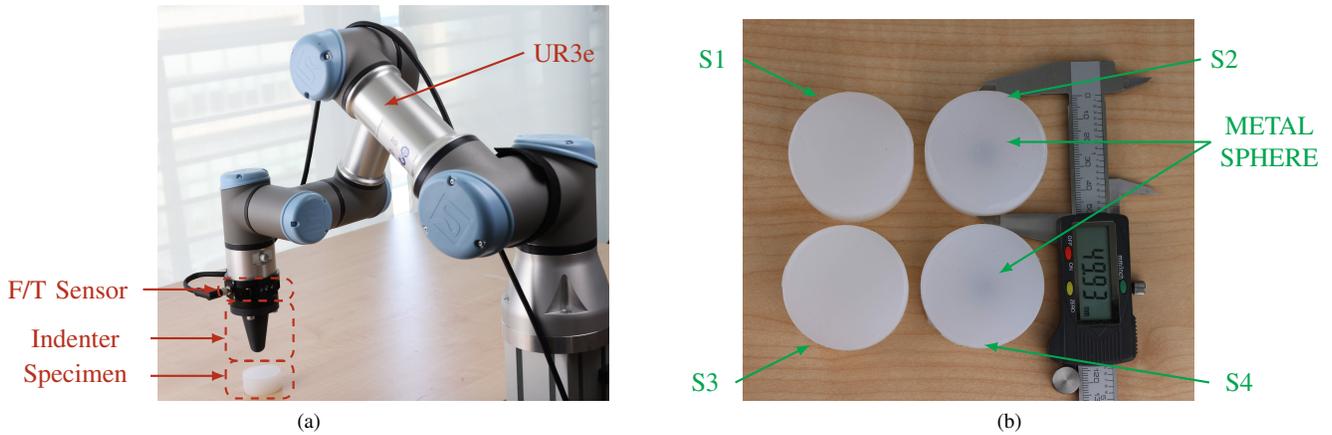

    \centering
\subfloat[\label{fig:ur3}]{\input{robot}}
    \hfill
  \subfloat[\label{fig:silicons}]{\input{silicones}}
  \caption{(a) The experimental setup is composed of a position-controlled Ur3e, a 6-axis force torque sensor BOTA SensorONe, a 3D printed indenter and a silicone specimen. (b) The 4 types of silicone. S1 is  softer (ECOFLEX-0030), S2 is the soft silicone with the steel ball, S3 refers to the stiffer silicone (Dragonskin-10kN) and S4 is the same silicone as S3 with the steel ball.}
\end{figure*}

The experiments used a
6-DoF position-controlled robotic arm, specifically the UR3e showed in
\autoref{fig:ur3}.
The robot's end effector was equipped with a force torque sensor and a
3D-printed indenter. The 6-axis Bota SensOne was used as force sensor
and the indenter's tip is shaped as a $\SI{2}{\centi \meter}$ diameter
spherical cap. The communication between the robot and control was
achieved using ROS2 Humble. A controller that simulates the behaviour
of a torque-controlled robot was used since the robot is
position-controlled and does not exposes any interface for joint torque
inputs~\cite{scherzinger2017forward}. The control
law~\eqref{eq:cartesian_impedance_simplified} has been simplified by
neglecting the noisy contribution of the acceleration, resulting in
\begin{equation}
    -
    \boldsymbol{D}_d \Dot{\boldsymbol{x}}_{ee} +
    \boldsymbol{K}_d\Tilde{\boldsymbol{x}}_{ee} \approx
    \boldsymbol{F}^{ext}_{ee} ,
    \label{eq:real_imp}
\end{equation}
therefore implying the adaptation of~\eqref{eq:im_sys}
and~\eqref{eq:im_sys_nl}.

Specimens were fabricated using two types of silicone with different
stiffness: Dragonskin-10kN, which is stiffer and similar to
muscles, and ECOFLEX-0030, which is softer and similar to
fat. The same moulding procedure was used to manufacture all
specimens. Firstly, the two liquid reacting components were mixed, and
then the mixture was put in a vacuum chamber for degassing. The mould
used was a cylinder with a diameter of $\SI{5}{\centi \meter}$ and a
height of $\SI{2.1}{\centi \meter}$. In addition to pure silicone
specimens, we created silicone specimens with a metal sphere inside
to simulate a tissue containing cancerous cells. The metal
sphere, with a diameter of $\SI{9}{\milli \meter}$, was inserted into
the liquid mixture after the degassing phase. Since cancerous tissue
can be up to 100 times stiffer than healthy tissue, we believe that
the use of a metallic inclusion provides a good benchmark for our
application~\cite{Nadan2007EUSLesions}.
%for our test we believe that inserting a metal ball can simulate quite closely what happens in biological tissue~\cite{Nadan2007EUSLesions}. 
In the manufactured specimens, the sphere sits at a distance of about $\SI{1}{\centi \meter}$ from the bottom surface. The four specimens used during the experiments are shown in \autoref{fig:silicons}.

% \begin{figure}
%     \centering
%     \input{silicones}
%     \caption{The 4 types of silicone. S1 is the softer one (ECOFLEX-0030), S2 is the softer silicone with the steel ball, S3 refers to the stiffer one (Dragonskin-10kN) and S4 is the same silicone as S3 with the steel ball.}
%     \label{fig:silicons}
% \end{figure}

In the following discussion of the experiments, some abbreviations have been used:
\begin{itemize}
    \item R refers to the reference values computed offline with a least square method;
    \item M$_1$ refers to the first model~\eqref{eq:state_space_NL_FT} where the KV and the F/T sensor are used;
    \item M$_2$ refers to the second model~\eqref{eq:im_sys} where the KV and the impedance control are used;
    \item M$_3$ refers to the third model~\eqref{eq:state_space_NL_FT} where the DR and the F/T sensor are used;
    \item M$_4$ refers to the fourth model~\eqref{eq:im_sys_nl} where the DR and the impedance control are used.
\end{itemize}
Silicone samples will be referred to with the 
%Also for the silicone during the discussion, we will refer to them with the same 
abbreviations defined in~\autoref{fig:silicons}, i.e., S1 to S4. To
ensure a fair comparison, all tests were conducted using the same
filter initial condition
$\hat{\boldsymbol{x}}_0 = \left[1,1,0,0\right]$ and the same initial
covariance matrix $\boldsymbol{P}_0$, equal to an identity matrix.

\subsection{Models Validation}
\begin{figure*}[t!]
    \centering
    \includegraphics[trim={1.5cm 0.15cm 1.5cm 0.2cm},clip,width=\textwidth,height=\textheight,keepaspectratio]{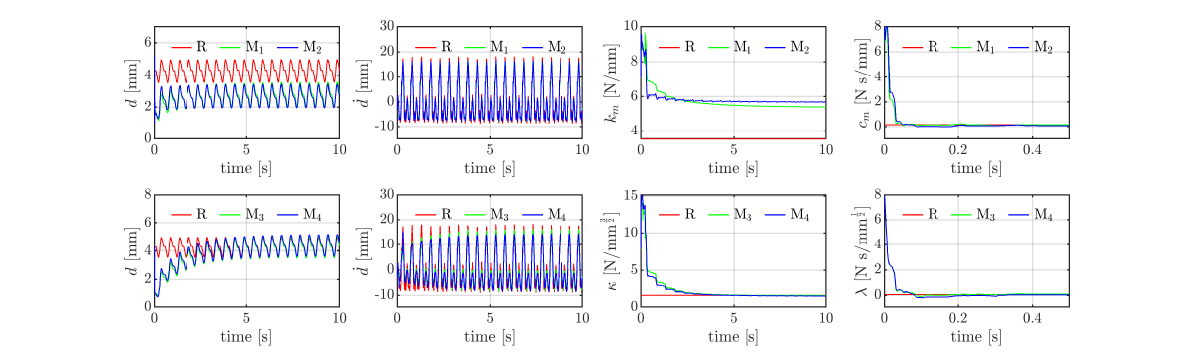}
    \caption{Results of the states estimation using the developed
      models with the softer silicone S1. In red with the label R are
      plotted the reference values computed with the LS method. In the
      first row are shown the results using the Kelvin-Voigt model
      with and without force sensor (M$_1$ and M$_2$). The second row
      shows the results of the estimation using the DM to model the
      soft body with and without the force torque sensor (M$_3$ and
      M$_4$).}
    \label{fig:models-comparison}
\end{figure*}

Figure~\ref{fig:models-comparison} shows the behaviour of the EKF with
the different types of models. We only report the results obtained for
the harder silicone sample, S1, as other samples behaviour and
thus considerations are reasonably similar. The reference
values are calculated by determining the precise location of the soft
body's surface and using the least squares method to identify the
stiffness and damping values that minimise the sum of the
residuals. The 4 plots in the first row refer to models M$_1$ and
M$_2$: upon examining the penetration, it is evident that the
estimated value of $x_1$ converges to a lower value than the actual
one, regardless of whether the F/T sensor is used. Additionally, the
stiffness value $x_3$ converges to a higher value than the actual
one. This discrepancy is due to the inaccurate description of the
system dynamics by the KV model, failing state values to converge to
the actual values.

The four plots in the second row refer to the models M$_3$ and M$_4$:
the penetration $x_1$ and the stiffness $x_3$ are now converging to
their expected values, as the velocity and viscosity do quite
rapidly. This is because the velocity is directly measured and
corrected within the EKF.

\subsection{Force Reconstruction}
As the identification of parameters can be performed without a force sensor, it is worth investigating how well the force can be reconstructed.
\begin{figure}[t!]
        \centering
\includegraphics[width=\columnwidth,height=\textheight,keepaspectratio]{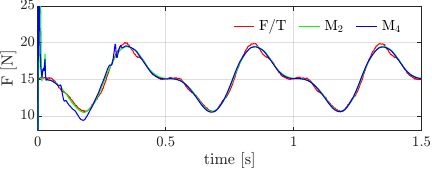}
        \caption{Comparison of the registered forces (F/T in red) against the estimated ones (M$_2$ in green and M$_4$) without the use of the sensor.}
        \label{fig:force}
\end{figure}
In~\autoref{fig:force} the comparison between the registered force and
the reconstructed forces is shown; again, the behaviour of only the
harder sample (S1) is shown. We can notice that after the initial
$\SI{30}{\milli \second}$ the force computed with M$_2$ and M$_4$ has an
absolute error of $\SI{0.4}{\newton}$. The mean squared error (MSE)
computed between $\SI{5}{\second}$ and $\SI{10}{\second}$, which is a
reasonable time for filter convergence, is equal to
$\SI{0.0614}{\newton^2}$ for M$_2$ and to $\SI{0.0559}{\newton^2}$ for
M$_4$. Despite converging to incorrect values, model M2 can precisely
track the force that is acting on the end effector. This demonstrates
the theoretical possibility of using the robot without a force sensor
and still accurately determining the force on the end effector. It is
worth noting that the estimated force is ahead of the measured
force. This is because the robot is not torque-controlled, but instead
uses a control that mimics the operation of an impedance
control~\cite{Schindeler2018OnlineLinearization}, thus inducing a
delay in the control chain.
%The control calculates new joint positions that minimise the difference between the force generated by the impedance checker and the force measured by the force sensor on the robot's end-effector.

\subsection{Tumour Identification}

\begin{table*}[t!]
\centering
\caption{Results of the estimation of $x_3$ and $x_4$ for each model obtained over the same dataset for each sample type after 5 and 10 seconds of palpation.} % All the filters are run over the same dataset for each sample type to compare the results. }
\label{tab:results_table}
%\resizebox{\columnwidth}{!}{%
\begin{tabular}{ccccccc}
\toprule %\cmidrule(l){3-12}
 \sc{Silicone} &  \sc{Model} & \sc{Time} & \multicolumn{2}{c}{\sc{Estimation}} &  \multicolumn{2}{c}{\sc{Reference}} \\
$[\#]$ & $[\#]$& $[\SI{}{\second}]$ & $x_3\;[\SI{}{\newton \milli\meter^{-1}}]$ & $x_4\;[\SI{}{\newton \second \milli\meter^{-1}}]$ &  $x_3\;[\SI{}{\newton \milli\meter^{-1}}]$ & $x_4\;[\SI{}{\newton \second \milli\meter^{-1}}]$\\
\midrule
S\textsubscript{1} & M\textsubscript{1} & $5 / 10$ & $3.45 / 3.41$ & $0.099 / 0.098$ & $2.03$ & $0.093$ \\
S\textsubscript{2} & M\textsubscript{1} & $5 / 10$ & $4.51 / 4.37$ & $0.120 / 0.118$ & $2.49$ & $0.118$ \\
S\textsubscript{3} & M\textsubscript{1} & $5/10$ & $5.49 / 5.38$ & $0.159 / 0.159$ & $3.53$ & $0.160$ \\
S\textsubscript{4} & M\textsubscript{1} & $5/10$ & $6.88 / 6.85$ & $0.119 / 0.119$ & $4.19$ & $0.121$ \\
S\textsubscript{1} & M\textsubscript{2} & $5/10$ & $3.39 / 3.31$ & $0.130 / 0.130$ & $2.03$ & $0.093$ \\
S\textsubscript{2} & M\textsubscript{2} & $5/10$ & $4.48 / 4.25$ & $0.118 / 0.122$ & $2.49$ & $0.118$ \\
S\textsubscript{3} & M\textsubscript{2} & $5/10$ & $5.33 / 5.29$ & $0.118 / 0.118$ & $3.53$ & $0.160$ \\
S\textsubscript{4} & M\textsubscript{2} & $5/10$ & $6.45 / 6.45$ & $0.025 / 0.026$ & $4.19$ & $0.121$ \\
\toprule
 \sc{Silicone} &  \sc{Model} & \sc{Time} & \multicolumn{2}{c}{\sc{Estimation}} &  \multicolumn{2}{c}{\sc{Reference}} \\
$[\#]$ & $[\#]$& $[\SI{}{\second}]$ & $x_3\;[\SI{}{\newton \milli\meter^{-\frac{3}{2}}}]$ & $x_4\;[\SI{}{\newton \second \milli\meter^{-\frac{1}{2}}}]$ &  $x_3\;[\SI{}{\newton \milli\meter^{-\frac{3}{2}}}]$ & $x_4\;[\SI{}{\newton \second \milli\meter^{-\frac{1}{2}}}]$\\
\midrule
S\textsubscript{1} & M\textsubscript{3} & $5/10$ & $0.961 / 0.877$ & $0.040 / 0.039$ & $0.742$ & $0.038$ \\
S\textsubscript{2} & M\textsubscript{3} & $5/10$ & $2.40 / 1.27$ & $0.047 / 0.053$ & $1.01$ & $0.052$ \\
S\textsubscript{3} & M\textsubscript{3} & $5/10$ & $1.76 / 1.70$ & $0.082 / 0.078$ & $1.70$ & $0.081$ \\
S\textsubscript{4} & M\textsubscript{3} & $5/10$ & $2.53 / 2.46$ & $0.068 / 0.066$ & $2.18$ & $0.069$  \\
S\textsubscript{1} & M\textsubscript{4} & $5/10$ & $0.856 / 0.791$ & $0.060 / 0.056$ & $0.742$ & $0.038$ \\
S\textsubscript{2} & M\textsubscript{4} & $5/10$ & $1.30 / 1.16$ & $0.069 / 0.064$ & $1.01$ & $0.052$ \\
S\textsubscript{3} & M\textsubscript{4} & $5/10$ & $1.69 / 1.61$ & $0.071 / 0.070$ & $1.70$ & $0.081$  \\
S\textsubscript{4} & M\textsubscript{4} & $5/10$ & $2.29 / 2.16$ & $0.044 / 0.048$ & $2.18$ & $0.069$ \\
\bottomrule
\end{tabular}
%}
\end{table*}

A sinusoidal palpation was performed for each silicone, whose results
are reported in~\autoref{tab:results_table}. As expected from
\autoref{fig:models-comparison}, the first two models,
M\textsubscript{1} and M\textsubscript{2}, converge to the wrong
stiffness and the correct damping value. The speed of convergence is
in the order of some seconds; in fact, the difference between the
values found at $\SI{5}{\second}$ and $\SI{10}{\second}$ is
minimal. Despite mismatches in the estimates, it is possible to
distinguish the silicones with and without the steel ball, so it is
theoretically possible to distinguish between diseased and healthy
tissues. Indeed, suppose the estimate of the diseased tissue is compared to
the estimate of the same healthy tissue in the vicinity. In that case, we can easily
recognise a difference in the stiffness value by comparison. Models
M$_3$ and M$_4$ on the other hand have slower convergence but they
converge to the correct value of stiffness and damping, thus ensuring
the detection of the silicone containing the metal ball. Models
M\textsubscript{2} and M\textsubscript{4}, which use the impedance
controller instead of the force sensor, tend to converge at lower
values of the stiffness. The reason for this mismatch is that the
impedance model within the EKF is just an approximation of the law that controls the robot. As explained before, it is due to the presence of a position control which only mimics an impedance behaviour. %there is a control loop that allows a position-controlled robot to be controlled with Cartesian impedance control. 
Despite this drawback, the different values still allows to distinguish the different materials.

\section{CONCLUSIONS}

This paper demonstrates the possibility of estimating soft tissue
characteristics and end effector penetration using dimensionality
reduction and an extended Kalman filter. The filter also allows for
the estimation of end effector forces without the need for additional
sensors. The experiments demonstrated the fast convergence of the filter and the accurate estimation of parameters for materials with different elasticity and viscosity, which makes it suitable for application which require rapid, but accurate, contact information. Additionally, hard intrusions, similar to cancerous cells, were positioned inside the soft material, to assess
the filter's ability to identify diseased tissues. Elements with
diameters lower than 1 cm were consistently found in multiple
experiments with different types of silicones.
  
Although we were able to estimate the desired properties, further improvements can be made. Firstly, a method to estimate the depth of hard intrusions would be beneficial for medical professionals. Secondly, a technique for detecting smaller intrusions should be developed. One potential solution to these issues is to merge the tactile estimation data with information gathered from an ultrasound scanner.


\balance
\bibliographystyle{IEEEtran}
\bibliography{references,ref2}

\end{document}